\documentclass[11pt,a4paper,logo]{googledeepmind}

\setleftlogo[180pt]{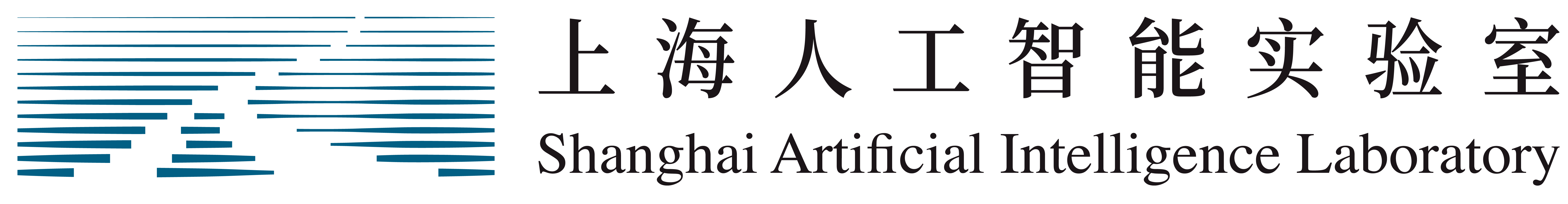}
\setrightlogo{}
\usepackage{fontawesome5}

\usepackage[T1]{fontenc}
\usepackage{pifont}
\usepackage[
    natbib=true,
    backend=biber,
    style=numeric, 
    sorting=none 
]{biblatex}
\AtEveryBibitem{\clearfield{month}}
\AtEveryBibitem{\clearfield{day}}
\usepackage{csquotes}

\newcommand{\sname}{MLEvolve\xspace}

\title{\sname: A Self-Evolving Framework for Automated Machine Learning Algorithm Discovery}
\homepageurl{https://internscience.github.io/MLEvolve/}
\codeurl{https://github.com/InternScience/MLEvolve}

\correspondingauthor{
$\spadesuit$: Co-first Authors \quad
$\clubsuit$: Corresponding Authors \\
Please send correspondence regarding this report to \href{mailto:yanxiangchao@pjlab.org.cn}{yanxiangchao@pjlab.org.cn} and \href{mailto:zhangbo@pjlab.org.cn}{zhangbo@pjlab.org.cn}.
}

\author[1,2,$\spadesuit$]{Shangheng~Du}
\author[1,$\spadesuit$,$\clubsuit$]{Xiangchao~Yan}
\author[1,2]{Jinxin~Shi}
\author[1]{Zongsheng~Cao}
\author[1]{Shiyang~Feng}
\author[1]{\protect\newline Zichen~Liang}
\author[1]{Boyuan~Sun}
\author[1]{Xu~Guo}
\author[1]{Tianshuo~Peng}
\author[1]{Yifan~Zhou}
\author[1]{Jie~Hou}
\author[1]{\protect\newline Xin~Li}
\author[1,2]{Jie~Zhou}
\author[1,2]{Liang~He}
\author[1,$\clubsuit$]{Bo~Zhang}
\author[1,$\clubsuit$]{Lei~Bai}

\affil[1]{Shanghai Artificial Intelligence Laboratory}
\affil[2]{East China Normal University}

\usepackage{pdflscape}

\usepackage{textcomp}
\usepackage{rotating}

\usepackage{setspace}
\usepackage{microtype} 

\usepackage{soul} 

\usepackage{graphicx}
\usepackage{subcaption}
\usepackage{caption}

\usepackage{booktabs}
\usepackage[table]{xcolor}
\sethlcolor{green!14}

\usepackage{array}
\usepackage{multirow}

\usepackage{amsmath}
\usepackage{siunitx}

\usepackage{enumitem}
\usepackage{float}
\usepackage{seqsplit}
\usepackage{framed}

\usepackage{tikz}
\usepackage{hyperref}
\usepackage{url}
\usepackage{listings}
\usepackage{xcolor}
\usepackage{soul}
\usepackage{colortbl}  
\usepackage{graphicx}
\usepackage{wrapfig}
\usepackage[most]{tcolorbox}
\usepackage{enumitem}
\usepackage{hyperref}

\tcbuselibrary{listingsutf8}
\definecolor{mycolor}{RGB}{50,80,150}

\usepackage{ragged2e}
\usepackage[most]{tcolorbox}

\usepackage{makecell}
\usepackage{adjustbox}

\usepackage[symbol]{footmisc}
\newcolumntype{Y}{>{\RaggedRight\arraybackslash}X}

\newcolumntype{C}{>{\centering\arraybackslash}X} %
\usepackage{colortbl}
\usepackage{xspace}
\usepackage{pifont}
\usepackage{bbding}
\usepackage{multirow}
\usepackage{booktabs,tabularx,xcolor}
\usepackage{enumitem}
\usepackage{ulem}
\usepackage{tabularx}
\usepackage{array}
\usepackage{wrapfig}
\usepackage{booktabs,tabularx,array}
\newcolumntype{L}{>{\raggedright\arraybackslash}X}
\newcolumntype{Y}{>{\centering\arraybackslash}X}  %
\newcolumntype{Z}{>{\raggedleft\arraybackslash}X}  %
\usepackage{textcomp}

\definecolor{oursbg}{RGB}{226,243,226}
\definecolor{ourshead}{RGB}{211,232,211}
\definecolor{groupbg}{RGB}{241,241,241}
\definecolor{summarybg}{RGB}{235,242,252}
\definecolor{lightgraytext}{RGB}{125,125,125}
\newcommand{\na}{\textcolor{lightgraytext}{--}}
\newcommand{\best}[1]{\textbf{#1}}
\newcommand{\second}[1]{\underline{#1}}
\newcommand{\ourscell}[1]{\cellcolor{oursbg}#1}

\newcommand{\grouprow}[1]{\rowcolor{groupbg}\multicolumn{8}{l}{\textbf{#1}}\\}
\usepackage{hyperref}
\hypersetup{
    colorlinks=true,
    linkcolor=blue, 
    citecolor=blue,  
    filecolor=black,
    urlcolor=blue    
}

\usepackage{tocloft}

\usepackage{etoolbox}
\usepackage{ulem} 
\makeatletter
\patchcmd{\@tocline}
    {\hfil}
    {\leaders\hbox{\hfil}\hfil}
    {}{}
\makeatother

\begin{document}
\sloppy

\begin{abstract}
Large language model (LLM) agents are increasingly applied to long-horizon tasks such as scientific discovery and machine learning engineering (MLE), where sustained self-evolution becomes a key capability. However, existing MLE agents suffer from isolated exploration, inefficient experience reuse, and lack of hierarchical control, which together hinder long-horizon optimization.
We present \sname, an LLM-based self-evolving multi-agent framework for end-to-end machine learning algorithm discovery. By extending tree search to Progressive MCGS, \sname enables cross-branch information flow through graph-based reference edges and gradually shifts the search from broad exploration to focused exploitation with an entropy-inspired progressive schedule. To allow the agent to evolve with accumulated experience, we introduce Retrospective Memory, which combines a cold-start domain knowledge base with a dynamic global memory for task-specific experience retrieval and reuse. For stable long-horizon iteration, we further decouple strategic planning from code generation with adaptive coding modes.
Evaluation on MLE-Bench shows that \sname achieves state-of-the-art performance across multiple dimensions including average medal rate and valid submission rate under a 12-hour budget (half the standard runtime). Moreover, \sname also outperforms specialized algorithm discovery methods including AlphaEvolve on mathematical algorithm optimization tasks, demonstrating strong cross-domain generalization.
\end{abstract}

\maketitle


\section{Introduction}

Artificial intelligence (AI) is reshaping scientific research and complex engineering, leading to the paradigm of AI for Science~\citep{van2023ai4Science}. With the continued advancement of large language models (LLMs), LLM-based agent systems~\citep{du2026survey} are now being applied to long-horizon autonomous tasks such as scientific discovery~\citep{aiscientist,team2025novelseek}, automated experimentation~\citep{feng2026internagent}, and end-to-end algorithm design~\citep{novikov2025alphaevolve}. Unlike single-turn reasoning, these scenarios involve open search spaces and limited time budgets, where agents must continually generate solutions, execute code, evaluate outcomes, and adjust strategies based on feedback. During this process, the agent continuously evolves: accumulating experience from past trials, adaptively adjusting exploration strategies, and progressively refining implementations according to the current search stage. This sustained self-evolving capability is becoming central to long-horizon autonomous agents.

Machine Learning Engineering (MLE) is one of the most representative scenarios for such long-horizon self-evolving tasks. Designing high-performance AI systems still relies heavily on expert knowledge and extensive manual iteration~\citep{amershi2019software-mle}. Although recent advances in AutoML~\citep{he2021automl,feurer2022auto-sklearn} have achieved significant progress in optimizing discrete stages such as data processing and model selection, they often fall short of covering the entire end-to-end MLE pipeline, \textit{i.e.}, from data preparation to model training and inference.
Recently, LLM-based coding agents have been applied to MLE scenarios~\citep{wang2024openhands,mlab,aide,rdagent,ml-master}, using the planning and code generation capabilities of LLMs to iteratively optimize within open search spaces. These agents typically employ greedy or evolutionary search~\citep{aide,li2025fm}, Monte Carlo Tree Search~\citep{ml-master,dojo}, or multi-agent collaboration~\citep{rdagent} to explore candidate solutions.

Despite these advances, existing MLE agents still face three key challenges that hinder self-evolution over long horizons.
First, existing search mechanisms suffer from \textbf{\textit{isolated exploration}} and lack adaptive exploration strategies. Most methods adopt linear or tree-structured search~\citep{aide,ml-master,dojo}, confining information within individual branches and making it difficult to transfer successful strategies across different search trajectories. Moreover, these methods generally employ fixed exploration strategies throughout the optimization process, leading to inefficient resource allocation under limited time budgets.
Second, most search frameworks suffer from \textbf{\textit{inefficient experience reuse}}, propagating only scalar rewards without reusing insights from earlier attempts~\citep{aide,ml-master,dojo}. While some recent methods explore memory mechanisms~\citep{automind,chen2026mars,zhu2026mlmaster2}, they require extra LLM calls or provide only static knowledge, lacking automatic experience accumulation during search.
Third, most existing methods couple planning and code implementation into one-shot generation, \textbf{\textit{lacking hierarchical control}}. A reasonable design requires distinguishing \emph{what to modify} from \emph{how to implement}, yet many methods~\citep{ml-master,du2025automlgen} rewrite the entire solution at every iteration, resulting in low iteration efficiency and uncontrollable modifications.

\begin{figure}[thbp]
\begin{center}
    \vspace{-0.5em}
    \includegraphics[width=0.96\linewidth]{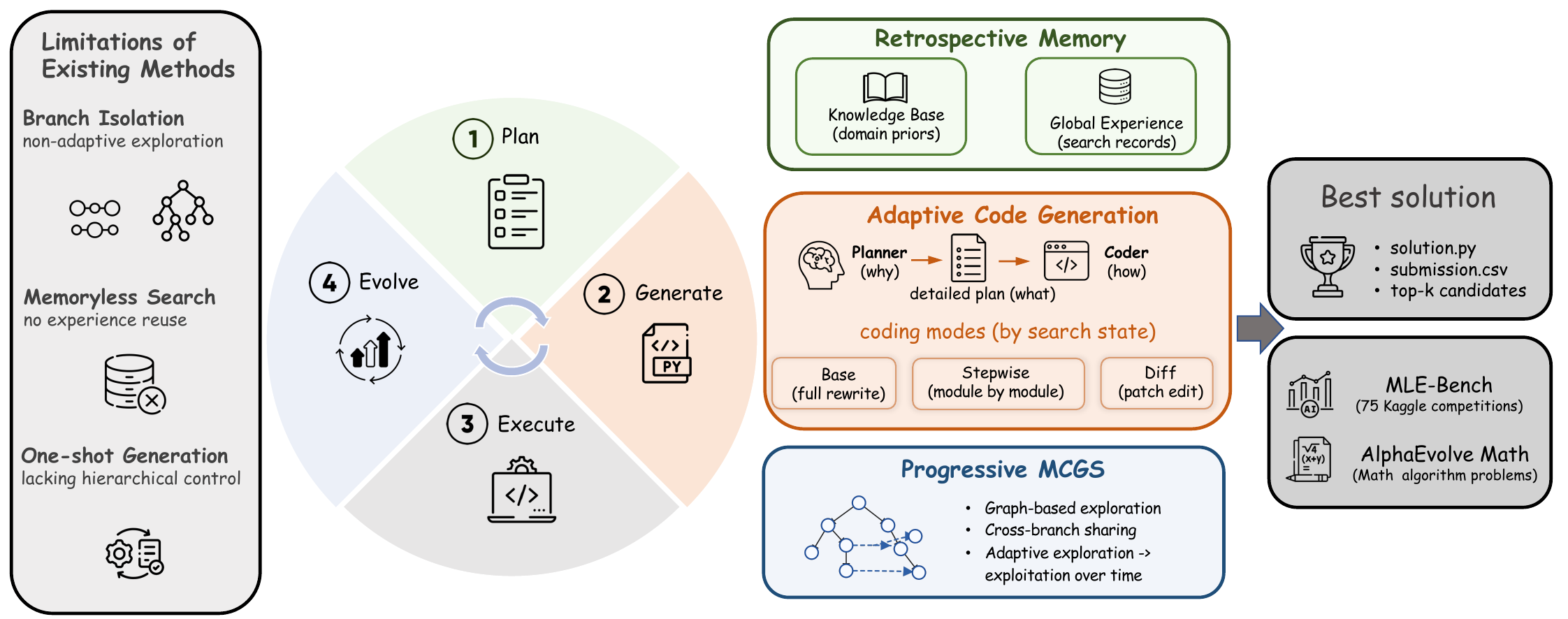}
    \vspace{-0.5em}
    \caption{\textbf{Overview of \sname that summarizes its core components and supported tasks.} Existing MLE agents suffer from isolated exploration, inefficient experience reuse, and lack of hierarchical control. \sname addresses these through Progressive MCGS, Retrospective Memory, and Hierarchical Planning with Adaptive Code Generation, supporting long-horizon iterative optimization tasks, such as end-to-end MLE and mathematical algorithm discovery.}
    \label{fig:overview}
\end{center}
\end{figure}

To bridge this gap, we present \textbf{\sname} (Figure~\ref{fig:overview}), an LLM-based self-evolving multi-agent framework for MLE tasks. \sname unifies three core components: (1)~\textbf{Progressive Monte Carlo Graph Search (MCGS)}, which addresses isolated exploration in tree search through graph-based cross-branch information flow, and introduces an entropy-inspired progressive exploration schedule that adaptively steers the search from broad exploration to focused exploitation over time; (2)~\textbf{Retrospective Memory}, pairing a curated domain knowledge base for cold-start initialization with a dynamic global memory that automatically accumulates and retrieves task-specific experience throughout the search; and (3)~\textbf{Hierarchical Planning with Adaptive Code Generation}, which separates strategic planning from code generation and selects among full rewrite, stepwise, and diff-based editing modes according to the current search state.
Consequently, \sname achieves more stable and self-evolving exploration of end-to-end ML pipelines, leading to stronger solutions for challenging MLE tasks. Experimental results show that \sname achieves a 65.3\% average medal rate on MLE-Bench under a 12-hour budget (half the standard runtime), establishing state-of-the-art performance, and further outperforms specialized algorithm discovery methods including AlphaEvolve~\citep{novikov2025alphaevolve} on mathematical optimization tasks.

Our key contributions are as follows:

\begin{itemize}[leftmargin=*]
    \item We propose \sname, a self-evolving multi-agent framework for end-to-end MLE tasks, which unifies progressive graph search, retrospective memory, and hierarchical adaptive code generation to support long-horizon iterative optimization.
    
    \item We introduce Progressive MCGS and Retrospective Memory for self-evolving optimization. Progressive MCGS resolves isolated exploration through graph-based cross-branch information flow and a progressive exploration schedule, while Retrospective Memory enables automatic experience accumulation and retrieval throughout the search.
    
    \item Extensive experiments show that \sname achieves a 65.3\% average medal rate on MLE-Bench under a 12-hour budget, achieving the best among all existing methods, and further outperforms AlphaEvolve~\citep{novikov2025alphaevolve} and AlphaEvolve-v2~\citep{alphaevolve_v2} on mathematical optimization tasks, demonstrating cross-domain generalization.

\end{itemize}


\section{Related Work}
\label{related_works}

\subsection{Automated Machine Learning Algorithm Discovery}
To address the unique challenges of MLE, a dedicated class of coding agents has been developed~\citep{aide,mle-star,mlzero,ml-master}, with many evaluated on benchmarks such as MLE-Bench~\citep{mle-bench}. These agents primarily frame the problem as a search for an optimal code-based solution. Early works like AIDE~\citep{aide} employ greedy search, which is susceptible to local optima. Subsequent frameworks adopt more structured exploration. ML-Master~\citep{ml-master} and AIRA-Dojo~\citep{dojo} use MCTS, MARS~\citep{chen2026mars} introduces budget-aware MCTS with contrastive reflection, and FM-Agent~\citep{li2025fm} applies evolutionary multi-island parallel search. Other works explore agent collaboration, such as R\&D-Agent~\citep{rdagent} with researcher-developer combination and AIBuildAI~\citep{zhang2026aibuildai} with hierarchical multi-agent coordination. Several methods also incorporate external knowledge or memory. AutoMind~\citep{automind} and Leeroo~\citep{nadafian2026kapso} ground search with domain knowledge bases, while ML-Master 2.0~\citep{zhu2026mlmaster2} introduces hierarchical cognitive caching for cross-task knowledge distillation. However, these methods commonly suffer from inter-branch information isolation and the inability to accumulate and reuse experience from past trials. Our method addresses these limitations from a self-evolving perspective, enabling the agent to continuously adapt its search behavior, accumulate experience, and refine solutions during long-horizon optimization.

\subsection{Graph-based Planning and Search}
Early methods that combine graph structures with MCTS, often referred to as MCGS~\citep{mcgs1,mcgs2}, were primarily developed for planning and reinforcement learning tasks with well-defined state spaces, where identical states are merged to compress the search space. Recent graph-based frameworks such as LocAgent~\citep{locagent2025} and CodexGraph~\citep{codexgraph2025} use graphs as static dependency representations for retrieval or localization, but these graphs do not evolve during search. In contrast, our MCGS targets open-ended LLM-based code generation, where each node represents a distinct candidate solution. The graph structure is not used to compress the state space, but to enable cross-branch information flow, trajectory reuse, and solution composition through dynamic reference edges.

\subsection{Memory and Experience Mechanisms for LLM Agents}
Memory mechanisms have been explored to improve LLM agent performance in iterative tasks~\citep{zhang2025memsurvey}. Recent work on long-term episodic memory~\citep{xu2026amem} enables agents to accumulate and retrieve experiential records across extended horizons, supporting more informed subsequent decisions. In the MLE domain, recent works further explore experience reuse. ROME~\citep{zhang2026rome} introduces ``reasoning gradients'' as structured optimization directions and stores successful trajectories as momentum memory, MARS~\citep{chen2026mars} extracts insights through contrastive reflection over historical attempts, and ML-Master 2.0~\citep{zhu2026mlmaster2} introduces hierarchical cognitive caching for cross-task knowledge distillation. While these methods advance experience reuse, most require additional LLMs for reflection or summarization. Our retrospective memory automatically accumulates and retrieves experience 
without requiring additional LLMs for explicit reflection, and further incorporates a static domain knowledge base for cold-start initialization.


\section{\sname}
\label{sec:method}

\begin{figure}[t]
\centering
\includegraphics[width=\linewidth]{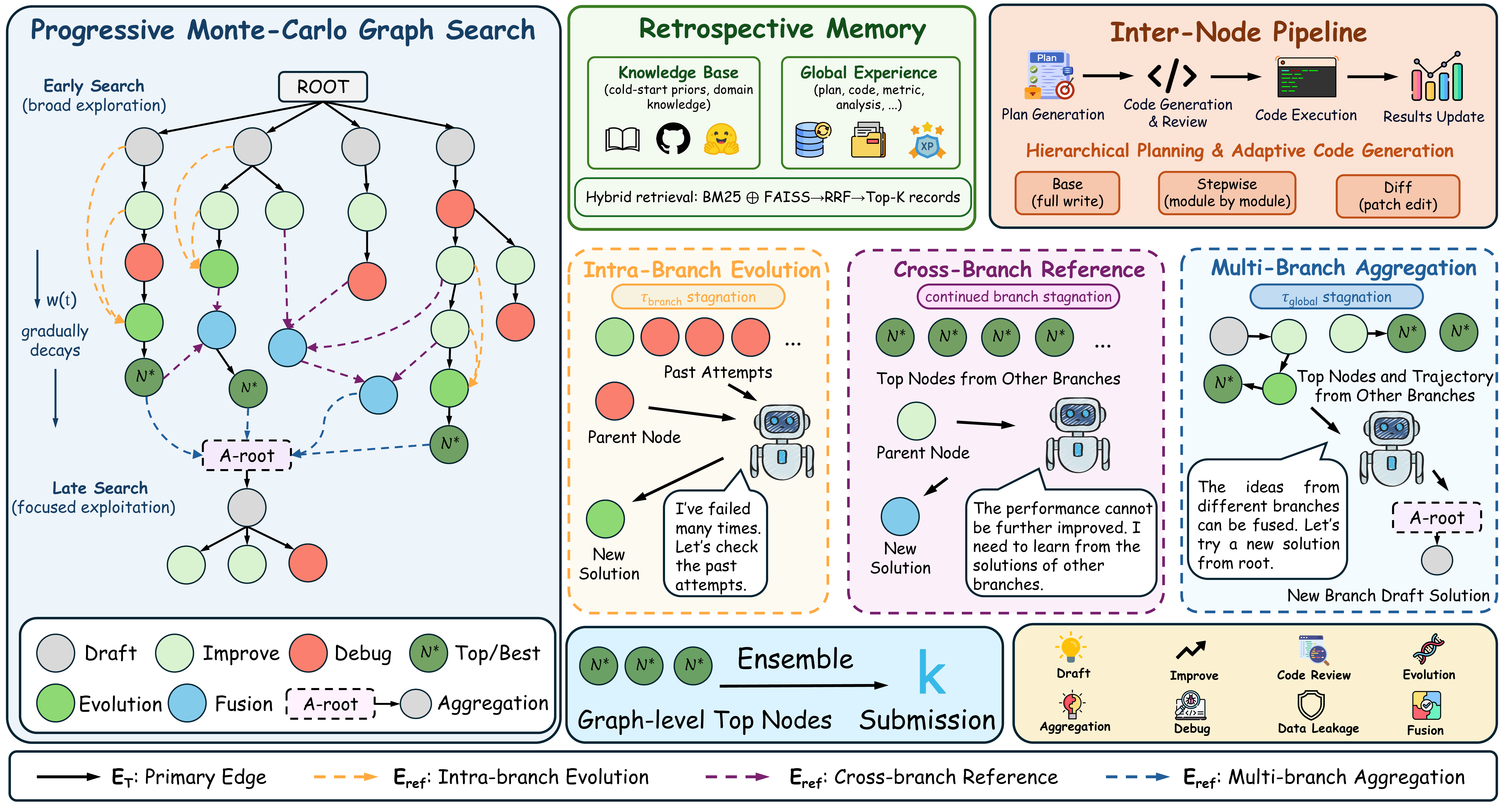}
\caption{\textbf{Framework of \sname.} The framework consists of three components. (i) Progressive MCGS extends MCTS with graph-based cross-branch information flow and a progressive exploration schedule. (ii) Retrospective Memory pairs a cold-start knowledge base with a dynamic global memory for experience accumulation and retrieval. (iii) Hierarchical Planning with Adaptive Code Generation decouples strategic planning from code implementation and selects among different coding modes according to the search state.}
\label{fig:framework}
\end{figure}

In automated algorithm discovery, strong solutions often arise from careful design, accumulated experience, and reference to multiple candidate pathways, rather than from a single linear refinement. To this end, we introduce \textbf{\sname}, a self-evolving multi-agent framework for MLE tasks. As shown in Figure~\ref{fig:framework}, the design combines three key components:
(1)~\textbf{Progressive MCGS} (\S\ref{sec:pmcgs}), which extends MCTS with graph-based cross-branch information sharing and a progressive exploration schedule to transition from broad exploration to focused exploitation;
(2)~\textbf{Retrospective Memory} (\S\ref{sec:memory}), combining a static domain knowledge base for cold-start initialization with a dynamic global memory that automatically accumulates and retrieves historical experience during search;
and (3)~\textbf{Hierarchical Planning with Adaptive Code Generation} (\S\ref{sec:codegen}), which separates strategic planning from code generation and selects different coding modes according to the current search state.


\subsection{Problem Formulation}

Our objective is to automate the search, design, and optimization of end-to-end ML pipelines. We formalize the task as identifying the optimal solution within a structured search space~\citep{aide}, where each node represents a complete candidate solution covering preprocessing, feature engineering, model training, and prediction.
The goal is to find the optimal solution for a given task:
\begin{equation}
\label{eq:objective}
s^{*} = \arg\max_{s \in \mathcal{S}} h(T, s),
\end{equation}
where $h(T,s)$ denotes the evaluation of candidate solution $s$ on task $T$, which may vary by task (\textit{e.g.}, accuracy, AUC, or loss). The solution space $\mathcal{S}$ is organized as a directed graph and explored through iterative search.

\subsection{Progressive MCGS}
\label{sec:pmcgs}

The search strategies in existing MLE methods face limitations such as branch information isolation and overly fixed search behavior. Greedy and evolutionary algorithms are prone to becoming trapped in local optima, while tree-search-based methods often spend substantial resources exploring low-value branches under limited time budgets, leading to inefficient resource allocation in later stages.
To address these limitations, we propose Progressive MCGS, which introduces a graph structure that enables cross-branch information sharing and a progressive exploration schedule that adaptively balances exploration and exploitation over time.

\subsubsection{Graph-based Search Space}
\label{sec:graph}

To realize the optimization objective in Eq.~(\ref{eq:objective}), we organize the search process as a directed graph:
\begin{equation}
\label{eq:graph}
G = (V, E), \quad E = E_{T} \cup E_{\text{ref}},
\end{equation}
where each node $v \in V$ maps to a candidate solution $s(v) \in \mathcal{S}$. Directed edges capture both generative and referential relationships:

\begin{itemize}[leftmargin=*]
    \item \textbf{Primary edges $E_{T}$}: $(u,v) \in E_{T}$ means that $v$ is derived from $u$ by applying an operator $o$, \textit{i.e.}, $v = g_{o}(u, R)$. These edges preserve the parent--child generative order and are used for selection and backpropagation.
    \item \textbf{Reference edges} $E_{\text{ref}}$: $(r,v) \in E_{\text{ref}}$ denotes that $v$ additionally incorporates information from node $r$ beyond its parent node. These edges connect nodes across branches or non-adjacent levels, enabling cross-branch knowledge flow and compositional transfer, but do not participate in backpropagation. When $E_{\text{ref}}=\varnothing$, the search reduces to standard MCTS.
\end{itemize}

\subsubsection{Progressive MCGS-based Exploration}
\label{sec:mcgs_exploration}

MCGS follows the MCTS loop of selection, expansion, simulation, and backpropagation, with progressive scheduling from broad exploration to elite-guided exploitation and graph-based expansion for cross-branch information reuse. We also provide a theoretical analysis in Appendix~\ref{appen:theory}.

\textbf{Selection with Progressive Exploration Scheduling.}
\label{sec:entropy}
Although the overall search space is formulated as a graph, UCT-based selection operates solely on the tree backbone formed by the primary edges $E_T$. When UCT exploration is selected, the selection policy traverses $E_T$ in a top-down manner to identify a node $v_t$ for expansion using the UCT criterion:
\begin{equation}
\pi_{\text{sel}}(v) = \arg\max_{i \in \mathcal{C}(v)} \text{UCT}(i),
\quad \text{where } \text{UCT}(i) = Q_i + c(t) \sqrt{\frac{\ln (N_v + 1)}{N_i + \varepsilon}},
\end{equation}
where $Q_i$ denotes the average reward of child node $i$, $N_i$ is its visit count, $N_v$ is the visit count of the parent, and $\varepsilon > 0$ is a smoothing constant. The exploration constant $c(t)$ is gradually reduced over time following a piecewise schedule ($c_0 \rightarrow c_{\min}$).

Inspired by entropy-based exploration principles~\citep{jaynes1957information}, we introduce an entropy-inspired progressive exploration schedule that transitions the search from broad exploration toward focused exploitation. Within a local time window, the branch selection frequencies form an empirical distribution $\pi_t$, whose Shannon entropy $H(\pi_t) = -\sum_i \pi_t(i)\log \pi_t(i)$ quantifies the dispersion of search effort. The core mechanism is a probabilistic soft switch between UCT-based exploration (higher entropy) and Elite-Guided exploitation (lower entropy). At each step, the system chooses between these strategies according to a time-dependent weight:
\begin{equation}
\label{eq:soft_switch}
P(S_t = \text{UCT}) = w(t), \qquad
P(S_t = \text{Elite}) = 1 - w(t),
\end{equation}
where $S_t$ denotes the selection strategy at step $t$, $w(t)$ gradually decreases from $1.0$ to a minimum threshold $w_{\min}$ as search time progresses. The schedule $w(t)$ is designed so that the empirical branch-selection entropy $H(\pi_t)$ progressively decreases over time, concentrating computation on promising branches. In the \textbf{Elite-Guided exploitation} mode, the system bypasses local tree traversal and selects from an elite set of top-$K$ globally best-performing nodes, weighted by inverse rank:
\begin{equation}
\label{eq:elite}
P(v_i \mid \text{elite set}) =
\frac{1/\text{rank}(v_i)}
{\sum_{j=1}^{K} 1/\text{rank}(v_j)},
\end{equation}
where $\text{rank}(v_i)$ is the position of node $v_i$ when all valid nodes are sorted by metric. This allows the search to directly exploit high-value nodes regardless of their position in the graph, while the probabilistic transition retains exploration capacity even in later stages.

\textbf{Expansion.}
To incorporate information flow and compositional reuse into the search process, we extend the standard MCTS expansion with graph-based operations. All expansion types are unified under a single formulation:
\begin{equation}
\label{eq:expansion}
v_{\text{new}} = g_{o}(v_{t}, R), \qquad (v_{t}, v_{\text{new}}) \in E_{T}, \;\; \{(r, v_{\text{new}}) \mid r \in R\} \subseteq E_{\text{ref}},
\end{equation}
where $R$ denotes the reference set. We instantiate this formulation with four expansion types (formal definitions in Appendix~\ref{appen:expansion}):

\textbf{(1) Primary expansion} ($R = \varnothing$).
The new node is generated solely from its parent without referencing other nodes. This constitutes the baseline expansion against which the graph-based variants extend.

\textbf{(2) Intra-branch evolution} ($R = \mathcal{R}_{\text{hist}}(v_t, k)$).
Inspired by human problem-solving strategies, this mode emphasizes reflecting on past attempts instead of blind trial and error. The agent takes the nearest $k$ nodes within the same branch to form a local trajectory as the reference set, reviewing which changes improved outcomes or caused failures. Through self-reflection, the agent reinforces effective patterns while avoiding repeated mistakes.

\textbf{(3) Cross-branch reference} ($R = \mathcal{R}_{\text{cross}}(N)$).
In ML competitions, contestants often draw inspiration from community-shared solutions when progress stalls. Similarly, when a branch shows signs of stagnation, MCGS selects the top-$N$ nodes across all evaluated branches as references, enabling the agent to draw on strong solutions discovered in other branches.

\textbf{(4) Multi-branch aggregation} ($R = \mathcal{R}_{\text{agg}}$).
For complex tasks, progress often requires synthesizing complementary insights from multiple strong solutions. This resembles a form of collective intelligence: trajectories from different branches are merged and fragments of useful insights are combined to spark novel directions. A new branch root is created beneath $v_0$, serving as a fresh starting point. Representative cases are provided in Appendix~\ref{appen:case}.

\textbf{Simulation.}
After generating a candidate $v_{\text{new}}$, its code is executed in an interpreter. The execution outputs are parsed to extract the task-specific metric and execution logs. An immediate reward $R(v)$ is designed to reflect execution validity and performance contribution:
\begin{equation}
\label{eq:reward}
R(v) =
\begin{cases}
-1, & \text{if execution fails or no valid metric is obtained} \\
1,  & \text{if execution succeeds but does not improve the branch best} \\
2,  & \text{if execution succeeds and refreshes the branch best metric}.
\end{cases}
\end{equation}
This structure distinguishes failed runs, feasible but non-improving attempts, and actual improvements, yielding stable credit assignment during MCGS.

\textbf{Backpropagation.}
After simulation, the reward $R(v)$ is propagated to the root only along primary edges $E_T$. Reference edges $E_{\text{ref}}$ are excluded because they represent auxiliary information reuse rather than parent--child generation, and therefore should not participate in credit assignment. For each ancestor node $u$ on the primary path, we update its visit count $N_u$ and cumulative reward $W_u$:
\begin{equation}
N_u \leftarrow N_u + 1,\qquad W_u \leftarrow W_u + R(v),
\end{equation}
and compute the average value estimate:
\begin{equation}
\label{eq:backprop}
Q_u = W_u / (N_u + \varepsilon).
\end{equation}

\textbf{Multi-Level Stagnation Detection.}
\label{sec:stagnation}
While the soft-switch schedule governs the global exploration-exploitation transition, the graph-based operators introduced above are triggered by explicit stagnation conditions to prevent branches from falling into unproductive loops:

\begin{itemize}[leftmargin=*]
    \item \textbf{Branch-level stagnation}: triggered when a branch produces $\tau_{\text{branch}}$ consecutive expansions without improving its best metric. The system first attempts intra-branch evolution; in later stages when other branches have accumulated strong solutions, cross-branch reference is further activated to incorporate external knowledge.
    \item \textbf{Global-level stagnation}: triggered when the global best metric has not improved for $\tau_{\text{global}}$ steps, activating multi-branch aggregation.
\end{itemize}

\subsection{Retrospective Memory}
\label{sec:memory}

To enable experience accumulation during search, we introduce a retrospective memory that retrieves relevant historical experience before each planning decision, transforming the search into experience-driven decision-making. The memory comprises a static domain knowledge base for cold-start initialization and a dynamic global memory for runtime experience accumulation.

\subsubsection{Domain Knowledge Base}
\label{sec:kb}

Effective ML solution design typically relies on domain priors and hands-on experience. LLM internal knowledge alone is often insufficient for specialized tasks, leading to a high rate of cold-start errors. To mitigate this, we curate a lightweight domain knowledge base of candidate models, organized by task type. For different task types (\textit{e.g.}, image classification, natural language processing, tabular regression), the knowledge base provides suitable models together with concise usage guidelines, synthesized from open-source repositories and competition platforms. Given a task $T$, the system retrieves relevant entries $R_{KB}(T)$ by matching the task description against domain keywords, treated as an optional signal during initial solution generation:
\begin{equation}
\label{eq:kb_init}
s_{\text{init}} = \mathrm{Init}(T, R_{KB}(T)),
\end{equation}
where $\mathrm{Init}(\cdot)$ denotes the initialization procedure that generates the first plan and code.

\subsubsection{Dynamic Global Memory}
\label{sec:global_memory}

During search, the global memory accumulates structured records after each valid node execution including the plan, outcome, analysis, and feedback signal.

\textbf{Hybrid retrieval.} Records are retrieved via a combination of lexical keyword matching and FAISS~\citep{johnson2019faiss}-based semantic search, fused through Reciprocal Rank Fusion (RRF):
\begin{equation}
\label{eq:rrf}
\text{score}(d) =
\alpha \cdot \frac{1}{k + r_{\text{lex}}(d)}
+ (1 - \alpha) \cdot \frac{1}{k + r_{\text{vec}}(d)},
\end{equation}
where $r_{\text{lex}}(d)$ and $r_{\text{vec}}(d)$ denote the ranks of record $d$ in the lexical and vector retrieval results, respectively; $k$ is a smoothing constant; and $\alpha$ balances the two signals.

\textbf{Stage-aware retrieval.}
Agents retrieve memory records with stage-specific queries and filters:
\begin{itemize}[leftmargin=*]
    \item \textbf{Planning stage}: After generating an initial free-text plan, the agent uses it as a query to retrieve relevant successful and failed experiences. These records guide the refinement of the plan into a structured module-level specification, helping the agent reuse effective strategies while avoiding previously unsuccessful directions.
    
    \item \textbf{Debugging stage}: When encountering an execution error, the agent uses the error message as a query to retrieve similar resolved errors from memory, providing helpful debug strategies.
\end{itemize}


\subsection{Hierarchical Planning and Adaptive Code Generation}
\label{sec:codegen}

To address the lack of hierarchical control in one-shot code generation, we introduce a hierarchical generation pipeline that decouples strategic planning from code implementation and adaptively selects among different code generation modes according to the current search state.

\subsubsection{Planner-Coder Decoupling}

We decouple strategic planning from code generation to separate global reasoning from local implementation. The planner operates at the module level, using execution feedback, branch trajectories, and retrieved memory to decide \textbf{what} to modify and \textbf{why}. The coder then implements the planned changes at the code level, focusing on \textbf{how} to realize the modification while preserving the existing code structure and working functions.

\subsubsection{Adaptive Code Generation Modes}

Rather than applying a single code generation mode, the coder applies three coding modes with different granularity, selected according to the current search state and task requirements:

\begin{itemize}[leftmargin=*]
    \item \textbf{Base mode}: Full code generation from scratch. This mode constructs a complete solution when no reliable solution is available, especially during initial drafting.

    \item \textbf{Stepwise mode}: Module-by-module generation following the planner's specification. This mode is used for complex tasks that require multi-stage pipelines, where decomposing the solution into modules helps reduce generation difficulty.

    \item \textbf{Diff mode}: Targeted diff edits on the existing code. When a working solution already exists, this mode enables localized refinements with more stable and controlled modifications.
\end{itemize}

The framework is realized through a team of specialized agents, each tailored to a specific search phase or operator type. Detailed agent descriptions are provided in Appendix~\ref{appen:agents}.

\section{Experiments}
\label{sec:experiments}

\subsection{Experiment Setup}
\label{sec:setup}

\textbf{Benchmarks.} We evaluate \sname on two benchmarks. The primary benchmark is MLE-Bench~\citep{mle-bench}, introduced by OpenAI for end-to-end machine learning engineering, comprising 75 Kaggle tasks across three complexity levels (low, medium, and high), with full details and evaluation metrics in Appendix~\ref{appen:benchmark}. To assess cross-domain generalization, we also use 15 open-ended mathematical optimization tasks from AlphaEvolve~\citep{novikov2025alphaevolve}.

\textbf{Implementation details.} We adopt Gemini-3.1-Pro-preview as the backbone LLM for all agents, with temperature set to 1.0. Each task is assigned a maximum of 500 expansion steps and a 12-hour runtime, executed on 21 vCPUs, 234\,GB of RAM, and a single NVIDIA H200 GPU. Full hyperparameter settings are listed in Appendix~\ref{appen:hyper}.

\textbf{Baselines.} We compare \sname with a series of MLE agents, including both proprietary and open-source agent frameworks. The proprietary methods include FM-Agent~\citep{li2025fm}, MLE-STAR-Pro-1.5~\citep{mle-star}, MARS~\citep{chen2026mars}, MARS+~\citep{chen2026mars}, and AIBuildAI~\citep{zhang2026aibuildai}. The open-source methods include AIDE~\citep{aide}, R\&D-Agent~\citep{rdagent}, ML-Master~\citep{ml-master}, AIRA-Dojo~\citep{dojo}, Leeroo~\citep{nadafian2026kapso}, and ML-Master 2.0~\citep{zhu2026mlmaster2}. The baseline results in Table~\ref{tab:main_res} are taken from the MLE-Bench leaderboard or the corresponding papers.

\subsection{Main Results}
\label{sec:main_results}

\begin{table*}[!t]
    \centering
    \caption{Main results on MLE-Bench (75 tasks, full set). Medal rates are reported across three complexity levels and overall, along with valid submission rate, above-median rate, and gold medal rate. Results are mean $\pm$ SEM over 3 seeds. We group methods by whether their code is publicly available. Best results are in \textbf{bold}; second best is \underline{underlined}.}
    \vspace{-0.5em}
    \label{tab:main_res}
    
    \setlength{\tabcolsep}{3.5pt}
    \renewcommand{\arraystretch}{1.05}
    
    \scriptsize
    \begin{tabularx}{\linewidth}{@{}lCCCCC|CCC@{}}
    \toprule
    & & \multicolumn{4}{c|}{\textbf{Medal rate by complexity}} & \multicolumn{3}{c}{\textbf{Other evaluation dimensions}} \\
    \cmidrule(lr){3-6} \cmidrule(lr){7-9}
    \textbf{Agent} & \textbf{\mbox{Time (h)}} & \textbf{\mbox{Low (\%)}} & \textbf{\mbox{Medium (\%)}} & \textbf{\mbox{High (\%)}} & \textbf{\mbox{All (\%)}} & \textbf{\mbox{Valid (\%)}} & \textbf{\mbox{Med+ (\%)}} & \textbf{\mbox{Gold (\%)}} \\
    \midrule
    
    \multicolumn{9}{c}{\textbf{\textit{Proprietary Methods}}} \\
    \midrule  
    
    \multicolumn{9}{@{}l}{\textbf{FM-Agent}~\citep{li2025fm}} \\
    Gemini-2.5-Pro & 24 & 62.1±1.5 & 36.8±1.5 & 33.3±0.0 & 43.6±0.9 & 96.9±1.2 & 51.6±1.2 & 22.7±0.8 \\
    \arrayrulecolor{black!30}\midrule
    
    \multicolumn{9}{@{}l}{\textbf{MLE-STAR-Pro-1.5}~\citep{mle-star}} \\
    Gemini-2.5-Pro & 24 & 68.2±2.6 & 34.2±1.5 & 33.3±0.0 & 44.0±1.3 & 93.8±0.4 & 52.9±1.6 & 19.1±1.8 \\
    \arrayrulecolor{black!30}\midrule
    
    \multicolumn{9}{@{}l}{\textbf{MARS}~\citep{chen2026mars}} \\
    Gemini-3-Pro-preview & 24 & 74.2±1.5 & 52.6±3.0 & 37.8±2.2 & 56.0±1.5 & \underline{98.7±0.0} & 65.8±1.6 & 31.1±0.4 \\
    \arrayrulecolor{black!30}\midrule
    
    \multicolumn{9}{@{}l}{\textbf{MARS+}~\citep{chen2026mars}} \\
    Gemini-3-Pro-preview & 24 & \underline{78.8±1.5} & 60.5±1.5 & \underline{44.4±2.2} & 62.7±0.8 & \textbf{100.0±0.0} & \underline{74.2±0.9} & \underline{33.8±0.4} \\
    \arrayrulecolor{black!30}\midrule
    
    \multicolumn{9}{@{}l}{\textbf{AIBuildAI}~\citep{zhang2026aibuildai}} \\
    Claude-Opus-4.6 & 24 & 77.3±0.0 & \underline{61.4±0.9} & \textbf{46.7±0.0} & \underline{63.1±0.4} & \textbf{100.0±0.0} & 71.1±1.2 & 25.8±0.4 \\
    
    \arrayrulecolor{black}\midrule
    \multicolumn{9}{c}{\textbf{\textit{Open-Source Methods}}} \\
    \midrule  
    
    \multicolumn{9}{@{}l}{\textbf{AIDE}~\citep{aide}} \\
    o1-preview & 24 & 35.9±1.9 & 8.5±0.4 & 11.7±1.3 & 17.1±0.6 & 82.8±1.1 & 29.4±1.3 & 9.4±0.8 \\
    \arrayrulecolor{black!30}\midrule
    
    \multicolumn{9}{@{}l}{\textbf{R\&D-Agent}~\citep{rdagent}} \\
    gpt-5 & 12 & 68.2±2.6 & 21.1±1.5 & 22.2±2.2 & 35.1±0.4 & 53.3±0.0 & 40.4±0.9 & 16.4±0.9 \\
    \arrayrulecolor{black!30}\midrule
    
    \multicolumn{9}{@{}l}{\textbf{ML-Master}~\citep{ml-master}} \\
    DeepSeek-R1 & 12 & 48.5±1.5 & 20.2±2.3 & 24.4±2.2 & 29.3±0.8 & 93.3±1.3 & 44.9±1.2 & 17.3±0.8 \\
    \arrayrulecolor{black!30}\midrule
    
    \multicolumn{9}{@{}l}{\textbf{AIRA-Dojo}~\citep{dojo}} \\
    o3 & 24 & 55.0±1.5 & 22.0±1.2 & 21.7±1.1 & 31.6±0.8 & 97.5±0.3 & 45.5±0.8 & 17.3±0.4 \\
    \arrayrulecolor{black!30}\midrule
    
    \multicolumn{9}{@{}l}{\textbf{Leeroo}~\citep{nadafian2026kapso}} \\
    Gemini-3-Pro-preview & 24 & 68.2±2.6 & 44.7±1.5 & 40.0±0.0 & 50.7±1.3 & 50.7±1.3 & 50.7±1.3 & 21.3±2.0 \\
    \arrayrulecolor{black!30}\midrule
    
    \multicolumn{9}{@{}l}{\textbf{ML-Master 2.0}~\citep{zhu2026mlmaster2}} \\
    DeepSeek-V3.2-Speciale & 24 & 75.8±1.5 & 50.9±3.5 & 42.2±2.2 & 56.4±2.5 & 95.6±1.2 & 63.1±1.2 & 19.6±0.9 \\
    \arrayrulecolor{black!30}\midrule
    
    \multicolumn{9}{@{}l}{\textbf{\sname\ (ours)}} \\
    \rowcolor[RGB]{222,236,215}
    Gemini-3.1-Pro-preview & \textbf{12} & \textbf{80.3±1.5} & \textbf{64.0±0.9} & \textbf{46.7±0.0} & \textbf{65.3±0.8} & \textbf{100.0±0.0} & \textbf{76.0±2.3} & \textbf{34.7±0.0} \\
    
    \arrayrulecolor{black}\bottomrule
    \end{tabularx}
    \end{table*}

\noindent\textbf{\sname achieves state-of-the-art performance on MLE-Bench.} As shown in Table~\ref{tab:main_res}, under a 12-hour budget, \sname attains an average medal rate of \textbf{65.3\%} and a gold medal rate of \textbf{34.7\%}, achieving the best overall performance among all compared MLE agents. The results are consistent across difficulty levels, with medal rates of 80.3\%, 64.0\%, and 46.7\% on low, medium, and high complexity tasks, respectively. In addition, \sname achieves a 100\% valid submission rate and a 76.0\% above-median rate, meaning that its submissions surpass the human median Kaggle score in more than three quarters of the tasks. \sname outperforms both open-source and proprietary baselines at half the standard 24-hour budget.

\begin{table}[t]
\centering
\scriptsize
\renewcommand{\arraystretch}{1.16}
\setlength{\tabcolsep}{3.8pt}
\caption{Comparison on \textbf{15 mathematical programming tasks} grouped by problem type. $\uparrow$ / $\downarrow$ means higher/lower is better. Values are displayed with task-dependent precision. Best results are in \textbf{bold}; second best is \underline{underlined}. ``--'' indicates the result is not reported.}
\label{tab:alphaevolve}
\resizebox{\linewidth}{!}{%
\begin{tabular}{p{5.2cm} c r r r r r >{\columncolor{oursbg}\raggedleft\arraybackslash}r}
\toprule
\multicolumn{1}{>{\columncolor{groupbg}}l}{\textbf{Problem}} &
\multicolumn{1}{>{\columncolor{groupbg}}c}{$\uparrow\!/\!\downarrow$} &
\multicolumn{1}{>{\columncolor{groupbg}}r}{\textbf{AlphaEvolve}} &
\multicolumn{1}{>{\columncolor{groupbg}}r}{\textbf{AlphaEvolve-v2}} &
\multicolumn{1}{>{\columncolor{groupbg}}r}{\textbf{SimpleTES}} &
\multicolumn{1}{>{\columncolor{groupbg}}r}{\textbf{TTT-Discover}} &
\multicolumn{1}{>{\columncolor{groupbg}}r}{\textbf{OpenEvolve}} &
\multicolumn{1}{>{\columncolor{ourshead}}c}{\textbf{\sname}} \\
\midrule
\grouprow{Geometric packing / regions}
Packing hexagons in hexagons & $\downarrow$ & \second{3.930092} & 3.931 & 3.931 & \na & \na & \ourscell{\best{3.9284759302}} \\
Circle packing in a square & $\uparrow$ & 2.6358627564 & \na & \second{2.635983} & \na & \na & \ourscell{\best{2.6359830395}} \\
Circle packing in a rectangle & $\uparrow$ & \second{2.3658321334} & \na & \na & \na & \na & \ourscell{\best{2.3658323759}} \\
Heilbronn convex regions & $\uparrow$ & \second{0.0309368890} & 0.0309 & \na & \na & \na & \ourscell{\best{0.0309372079}} \\
Heilbronn triangles & $\uparrow$ & \second{0.03652988988003016} & 0.0365 & \na & \na & \na & \ourscell{\best{0.03652988988003020}} \\
Kissing number dimension 11 & $\uparrow$ & \best{593} & \best{593} & \na & \na & \na & \ourscell{592} \\
\midrule
\grouprow{Additive combinatorics}
Sums differences problems 1 & $\uparrow$ & \second{1.1479889651} & 1.1479 & 1.143975 & \na & \na & \ourscell{\best{1.1901774219}} \\
Sums and differences problems 2 & $\uparrow$ & \second{1.1584172816} & 1.1584 & \na & \na & \na & \ourscell{\best{1.1585457700}} \\
\midrule
\grouprow{Autocorrelation / inequalities}
An uncertainty inequality & $\downarrow$ & \second{0.3520991044225} & 0.3521 & \na & \na & \na & \ourscell{\best{0.3520991044160}} \\
First autocorrelation inequality & $\downarrow$ & 1.5052939684 & 1.5032 & 1.503871 & \second{1.5028628983} & 1.507190 & \ourscell{\best{1.5028628749}} \\
Third autocorrelation inequality variant & $\downarrow$ & \second{1.4687620697} & \na & \na & \na & \na & \ourscell{\best{1.4587698922}} \\
Third autocorrelation inequality & $\downarrow$ & 1.4556427954 & 1.4557 & \best{1.453675} & \na & 1.460000 & \ourscell{\second{1.4548507482}} \\
Second autocorrelation inequality & $\uparrow$ & 0.8962799442 & \second{0.961} & \best{0.962694} & 0.959100 & 0.944900 & \ourscell{0.9054217971} \\
\midrule
\grouprow{Ratio / overlap optimization}
Max-to-min ratios & $\downarrow$ & 12.88926611203 & \second{12.889266112} & \na & \na & \na & \ourscell{\best{12.8892299077}} \\
Minimum Overlap Problem & $\downarrow$ & 0.3809230351 & 0.380924 & \best{0.380868} & \second{0.3808753232} & 0.380965 & \ourscell{0.3808968496} \\
\bottomrule
\end{tabular}%
}
\end{table}

\noindent\textbf{\sname generalizes to mathematical algorithm optimization.} To further evaluate the generalization ability of \sname in self-evolving optimization scenarios, we apply it to the 15 mathematical optimization tasks from AlphaEvolve. These tasks differ from end-to-end MLE pipelines, but share a similar iterative optimization structure: the agent repeatedly proposes candidate solutions, evaluates their quality, and refines them through continued search.
As shown in Table~\ref{tab:alphaevolve}, \sname achieves the best result on \textbf{11 of 15} tasks when compared with specialized algorithmic discovery methods, including AlphaEvolve~\citep{novikov2025alphaevolve}, AlphaEvolve-v2~\citep{alphaevolve_v2}, SimpleTES~\citep{simpleTes}, TTT-Discover~\citep{ttt-discover}, and OpenEvolve~\citep{openevolve}. These results suggest that our self-evolving mechanism is not limited to the MLE domain, but can generalize to broader algorithmic optimization problems that require iterative optimization.

\subsection{Ablation Study}
\label{sec:ablation}

\begin{table}[t]
\centering
\caption{Component-level ablation on MLE-Bench Lite (22 tasks). Each row removes one core component of \sname. Beat Ratio is the average percentage of human Kaggle competitors outperformed. Best performances are marked in bold.}
\vspace{-0.5em}
\label{tab:ablation}
\small
\setlength{\tabcolsep}{6pt}
\renewcommand{\arraystretch}{1.05}
\begin{tabular}{lccc}
\toprule
\textbf{Configuration} & \textbf{Medal (\%)} & \textbf{Gold (\%)} & \textbf{Beat Ratio (\%)} \\
\midrule
\rowcolor[RGB]{222,236,215}
\textbf{\sname} & \textbf{81.82} & \textbf{54.55} & \textbf{88.39} \\
\quad w/o Progressive MCGS & 68.18 & 40.91 & 79.91 \\
\quad w/o Retrospective Memory & 68.18 & 50.00 & 81.90 \\
\quad w/o Adaptive Code Generation & 72.73 & 40.91 & 84.14 \\
\bottomrule
\end{tabular}
\end{table}

To evaluate the effectiveness of proposed components, we conduct ablation experiments on MLE-Bench Lite (22 tasks) by removing one component at a time while keeping all others unchanged.
As shown in Table~\ref{tab:ablation}, removing any single component leads to a clear performance decline, indicating that all three components help alleviate the existing limitations. Specifically, removing Progressive MCGS causes the largest drop in beat ratio and reduces the medal rate from 81.82\% to 68.18\%. Without this module, the search reverts to standard tree-based MCTS with a fixed strategy that wastes resources on low-value branches in later stages. Removing Retrospective Memory yields the same medal-rate decrease, highlighting the value of experience guidance. In this setting, the agent can still occasionally discover strong solutions through search alone, but it lacks experience feedback and guidance in long-horizon tasks. Replacing adaptive code generation with one-shot generation similarly reduces overall performance, since the absence of planner-coder decoupling and diff-based editing weakens the stability of iterative code refinement.
We further analyze the individual mechanisms within Progressive MCGS and Retrospective Memory in Appendix~\ref{appen:detailed_ablation}. The results show that intra-branch evolution is the most critical factor within Progressive MCGS, while Elite-Guided exploitation mainly improves leaderboard ranking by further refining already competitive solutions toward higher-performing ones.

\subsection{Further Analysis}
\label{sec:analysis}

\subsubsection{Progressive Search Entropy Dynamics}

\begin{figure}[t]
\centering
\includegraphics[width=0.6\linewidth]{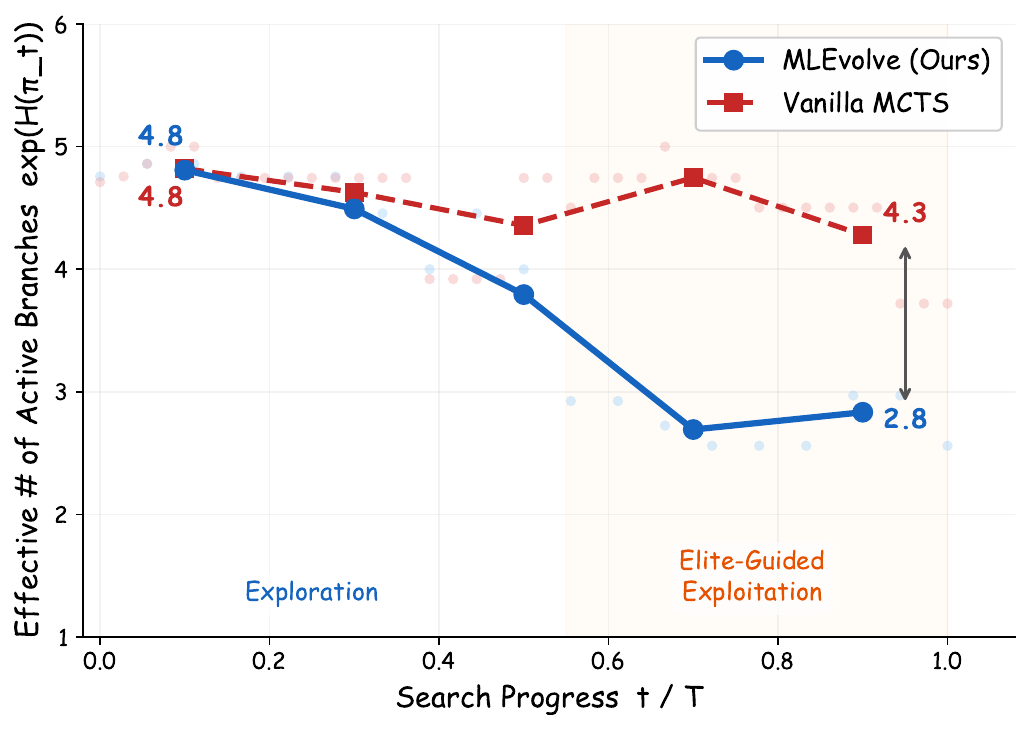}
\caption{Effective branch count $\exp(H(\pi_t))$ over search progress. \sname progressively reduces from 4.8 to 2.8, empirically validating the soft-switch schedule (Eq.~(\ref{eq:soft_switch})). Vanilla MCTS with a fixed exploration constant remains near 4.3 throughout.}
\label{fig:entropy}
\end{figure}

To empirically validate the progressive transition from exploration to exploitation described in \S\ref{sec:entropy}, we measure the effective number of active branches during search. Specifically, within a sliding window at search progress $t$, we compute the empirical distribution $\pi_t$ of branches selected for solution iteration, and use $\exp(H(\pi_t))$ to quantify the number of branches over which search effort is effectively distributed, where $H(\pi_t)$ is the Shannon entropy of $\pi_t$. 
As shown in Figure~\ref{fig:entropy}, \sname gradually reduces the effective number of active branches from 4.8 in the early exploration stage to 2.8 in the later exploitation stage, indicating that the soft switch schedule progressively concentrates computation on more promising candidates. In contrast, Vanilla MCTS remains almost uniform throughout, continuing to spread resources across branches even after promising directions emerge. The observed entropy trend is consistent with the scheduling behavior in Eq.~(\ref{eq:soft_switch}), showing the effectiveness of the progressive exploration schedule and Elite-Guided exploitation.

\subsubsection{Performance with Different LLMs}
\label{sec:multi_model}

\begin{figure}[t]
\centering
\includegraphics[width=0.6\linewidth]{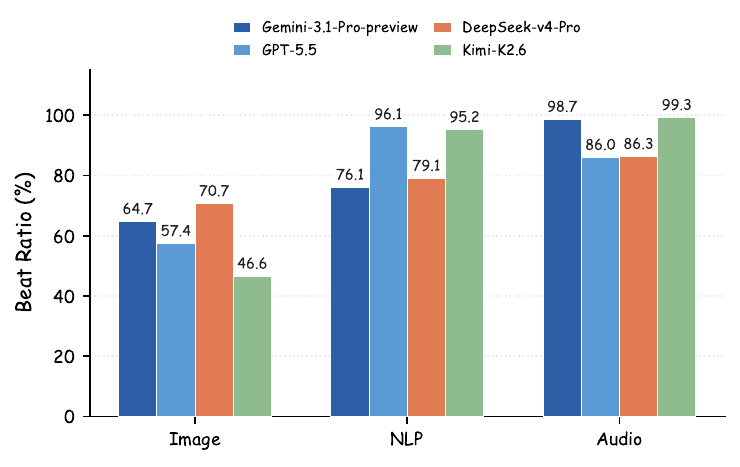}
\caption{Performance of \sname across different backbone LLMs on representative tasks covering Image, NLP, and Audio domains. Beat ratios are reported per domain, with full per-task scores provided in Appendix.}
\label{fig:multi_model}
\end{figure}

We further evaluate \sname with four LLM backbones, including Gemini-3.1-Pro-preview, GPT-5.5, DeepSeek-v4-Pro, and Kimi-K2.6, on representative MLE-Bench tasks covering Image, NLP, and Audio domains. As shown in Figure~\ref{fig:multi_model}, the four backbones exhibit clearly different domain strengths. For example, DeepSeek-v4-Pro leads on the evaluated Image tasks with 70.7\%, GPT-5.5 on NLP with 96.1\%, and Kimi-K2.6 on Audio with 99.3\%. Despite these per-domain differences, all four LLMs achieve competitive results under the same \sname pipeline, suggesting that the framework is not tightly coupled to a specific LLM backbone. These results show that \sname remains effective across different backbone LLMs and task domains. Full per-task scores are provided in Appendix~\ref{appen:multi_model}.

\subsubsection{Performance Over Time}

\begin{figure}[t]
\centering
\includegraphics[width=0.6\textwidth]{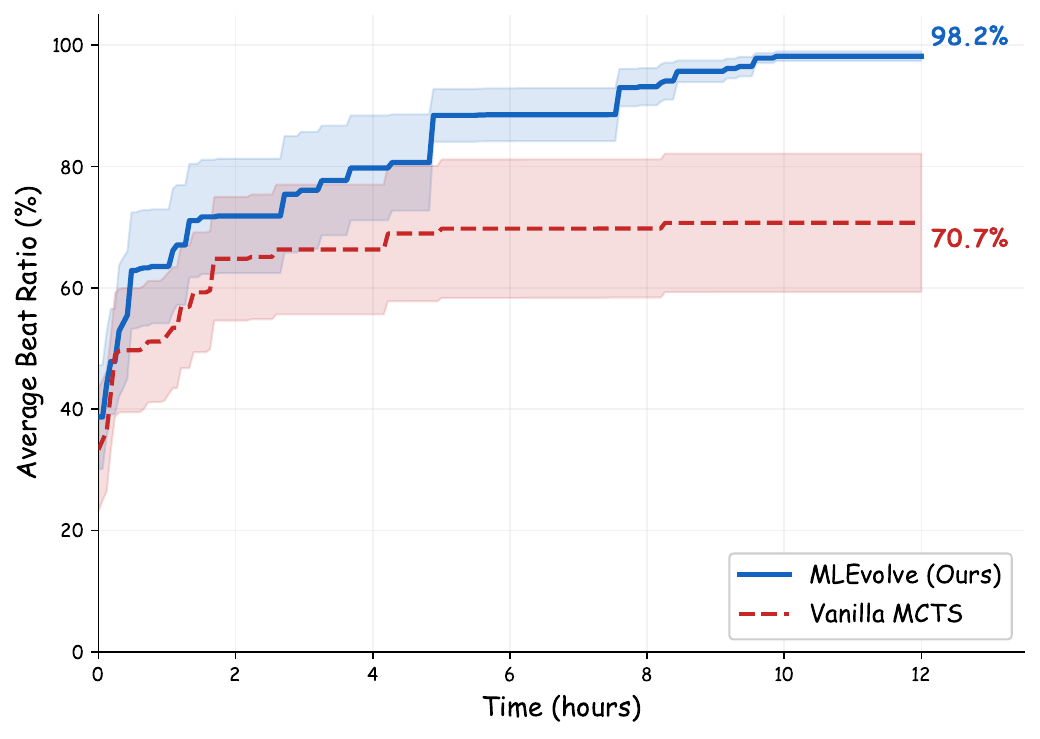}
\caption{Beat ratio over the 12-hour search budget on representative tasks. \sname converges faster and continues improving in late stages, whereas the baseline plateaus early.}
\label{fig:convergence}
\end{figure}

To examine how performance evolves during the search, Figure~\ref{fig:convergence} reports the beat ratio (the percentage of human Kaggle participants outperformed by the current best submission) as a function of elapsed time. \sname improves rapidly in the early stage and continues to make gains throughout the middle and late stages, reaching a final beat ratio of 98.2\% on the representative tasks. In contrast, Vanilla MCTS plateaus much earlier, ending at $\sim$70\% and struggling to further refine solutions once the early promising directions have been explored. This trend indicates that \sname can sustain improvement over a longer search horizon, supporting the effectiveness of the self-evolving design.


\section{Conclusion}
\label{sec:conclusion}

In this work, we present \sname, an LLM-based self-evolving multi-agent framework for long-horizon MLE tasks. By integrating Progressive MCGS, Retrospective Memory, and Hierarchical Planning with Adaptive Code Generation, \sname enables adaptive search, sustained experience accumulation, and flexible code generation within a unified optimization process. Experiments show that \sname achieves state-of-the-art performance on MLE-Bench, attaining a 65.3\% average medal rate under a 12-hour budget and outperforming all existing baselines. Ablation studies verify the effectiveness of each component. Results on AlphaEvolve mathematical optimization tasks further show that \sname generalizes beyond MLE to broader algorithmic optimization problems. In future work, we will extend \sname to more general AI for Science scenarios, including automated scientific experimentation, cross-disciplinary algorithm discovery, and autonomous research workflows.

\begingroup
\sloppy
\normalem
\printbibliography[heading=bibintoc]
\endgroup

\clearpage
\appendix

\newpage
\section*{\centerline{Appendix}}
\vspace{0.5em}

\section{Agent Descriptions}
\label{appen:agents}

\sname is realized through a team of specialized agents, each tailored to a specific search phase or operator type. We summarize their roles:

\begin{itemize}[leftmargin=*]
  \item \textbf{Draft Agent.} Generates initial candidate solutions at the root node, which can retrieve model priors from the cold-start knowledge base (\S\ref{sec:kb}).
  \item \textbf{Improve Agent.} Iteratively refines a runnable solution. It usually obtains structured guidance from the planner and applies controlled revisions through the Diff mode.
  \item \textbf{Debug Agent.} This agent is triggered only when execution fails. It repairs faulty solutions based on error traces (\textit{e.g.}, missing dependencies, tensor shape mismatches), applying minimal modifications until the issue is fixed or the retry limit is reached.
  \item \textbf{Evolution Agent.} Corresponds to intra-branch evolution by aggregating recent consecutive nodes along the same branch. It extracts experience from the past trajectory and uses it to propose targeted refinements for the current solution.
  \item \textbf{Fusion Agent.} Performs cross-branch reference when a branch stagnates. It aggregates strong solutions from other branches as additional references, supplying reusable strategies for the current solution.
  \item \textbf{Aggregation Agent.} Triggered by global stagnation, it aggregates top trajectories from multiple branches to create a new branch starting point.
  \item \textbf{Code Review Agent.} After each code generation step, it reviews the generated code for naming or import errors, suspicious patterns, and metric consistency before execution.
  \item \textbf{Data Leakage Agent.} Checks for potential leakage between training and evaluation splits to prevent overfitting to evaluation artifacts and avoid inflated scores.
  \item \textbf{Result Parse Agent.} Parses execution logs to extract task-specific metrics, execution status, and key insights, and transfers structured information back into the search loop.
\end{itemize}

\section{Expansion Type Formulations}
\label{appen:expansion}

In \S\ref{sec:mcgs_exploration} we introduce a unified expansion rule
\begin{equation*}
v_{\text{new}} = g_{o}(v_{t}, R), \qquad (v_{t}, v_{\text{new}}) \in E_{T}, \;\; \{(r, v_{\text{new}}) \mid r \in R\} \subseteq E_{\text{ref}},
\end{equation*}
parametrized by the reference set $R$. This appendix gives the precise instantiation of $R$ for each of the four expansion types.

\textbf{(1) Primary expansion} ($R = \varnothing$).
The new node is generated solely from its parent, without referencing other nodes:
\begin{equation}
v_{\text{new}} = g_{o}(v_{t}, \varnothing), \qquad (v_{t}, v_{\text{new}}) \in E_{T}.
\end{equation}
This is the baseline expansion against which the graph-based variants extend, and corresponds to operators such as \textit{Draft}, \textit{Improve}, and \textit{Debug}.

\textbf{(2) Intra-branch evolution} ($R = \mathcal{R}_{\text{hist}}(v_t, k)$).
The reference set consists of the nearest $k$ ancestor nodes of $v_t$ within the same branch, forming a local trajectory:
\begin{equation}
v_{\text{new}} = g_{o}(v_{t}, \mathcal{R}_{\text{hist}}(v_{t}, k)), \qquad (v_{t}, v_{\text{new}}) \in E_{T}, \;\; \{(r, v_{\text{new}}) \mid r \in \mathcal{R}_{\text{hist}}(v_t, k)\} \subseteq E_{\text{ref}}.
\end{equation}
The primary edge preserves the parent--child relation, while the reference edges record information flow from intra-branch history. Selection and backpropagation are conducted exclusively along $E_T$.

\textbf{(3) Cross-branch reference} ($R = \mathcal{R}_{\text{cross}}(N)$).
When the current branch stagnates, the system constructs a reference set from the top-$N$ nodes selected across evaluated branches according to their performance:
\begin{equation}
v_{\text{new}} = g_{o}(v_{t}, \mathcal{R}_{\text{cross}}(N)), \qquad (v_{t}, v_{\text{new}}) \in E_{T}, \;\; \{(r, v_{\text{new}}) \mid r \in \mathcal{R}_{\text{cross}}(N)\} \subseteq E_{\text{ref}}.
\end{equation}
These reference edges allow the new node to reuse effective designs discovered in other branches, providing external guidance for improving the current solution.

\textbf{(4) Multi-branch aggregation} ($R = \mathcal{R}_{\text{agg}}$).
Triggered by global stagnation, this operator creates a new branch beneath the root $v_0$ by aggregating top trajectories from multiple branches. Let $\mathcal{T}^{\text{top}}_{b}$ denote the best-performing trajectories in branch $b \in \mathcal{B}$; then $\mathcal{R}_{\text{agg}} = \bigcup_{b \in \mathcal{B}} \mathcal{T}^{\text{top}}_{b}$, and the expansion is:
\begin{equation}
v_{\text{new}} = g_{o}(v_{0}, \mathcal{R}_{\text{agg}}), \qquad (v_{0}, v_{\text{new}}) \in E_{T}, \;\; \{(u, v_{\text{new}}) \mid u \in \mathcal{R}_{\text{agg}}\} \subseteq E_{\text{ref}}.
\end{equation}
Unlike incremental refinement along a single branch, aggregation pools information from multiple branches and opens an independent exploration trajectory.

\section{More Theoretical Analysis}
\label{appen:theory}
This section states two properties of Progressive MCGS. The first characterizes the branch coverage created by Eq.~(\ref{eq:expansion}); the second characterizes the reuse concentration and top-$m$ sample count induced by elite-guided selection in Eq.~(\ref{eq:elite}).

For a node $z$ created by Eq.~(\ref{eq:expansion}), define its source set as $S(z)=\{p_T(z)\}\cup R(z)$, where $p_T(z)$ is the primary parent and $R(z)$ is the reference set supplied to the expansion. Let $b(v)$ denote the primary branch containing node $v$, and let $B(z)=\{b(u):u\in S(z)\}$. Construct an undirected branch-exposure graph $H_t=(B_t,L_t)$, where $\{i,j\}\in L_t$ if some expansion context contains at least one node from branch $i$ and at least one node from branch $j$. Reward backup remains on the primary path as defined in Eq.~(\ref{eq:backprop}).

\noindent\textbf{Proposition 1 (Coverage compactness).} If $B(z)$ contains $s\ge2$ branches, then those branches form a clique $K_s$ in $H_t$, covering $\binom{s}{2}$ branch-exposure edges. Therefore, an aggregation containing all $q$ target branches realizes the clique $K_q$ in one expansion. If instead each expansion context contains at most $r$ branches with $2\le r<q$, then at least
$\left\lceil\frac{\binom{q}{2}}{\binom{r}{2}}\right\rceil$
such expansions are required to cover the same $K_q$ branch-exposure edges.

\medskip\noindent\textit{Proof.} For every pair of branches $i,j\in B(z)$, the definition of $H_t$ includes edge $\{i,j\}$. Hence $B(z)$ becomes a clique $K_s$ with $\binom{s}{2}$ branch-exposure edges; when the aggregation contains all $q$ target branches, it forms $K_q$ with $\binom{q}{2}$ edges. A context containing at most $r$ branches can add at most $\binom{r}{2}$ such edges. Therefore, covering those $\binom{q}{2}$ edges requires at least $\lceil\binom{q}{2}/\binom{r}{2}\rceil$ contexts of width at most $r$.
\hfill\rule{0.35em}{0.35em}

\medskip\noindent Multi-branch aggregation is therefore compact in terms of context coverage: one $q$-branch context can create the same branch-exposure clique that would otherwise require multiple narrower reference contexts.

\noindent\textbf{Proposition 2 (Reuse concentration and sample count).} Fix an elite set of $K>1$ valid nodes, rank them from best to worst by the task metric, and break ties by a fixed deterministic rule. Let $H_m=\sum_{i=1}^m 1/i$, $p_{\mathrm{elite}}=H_m/H_K$, and $p_{\mathrm{uniform}}=m/K$. Under independent selections from this fixed set, Eq.~(\ref{eq:elite}) gives $p_i=(1/i)/H_K$; hence $p_{\mathrm{elite}}>p_{\mathrm{uniform}}$ for every $1\le m<K$, $\mathrm{E}[\mathrm{rank}]=\sum_{i=1}^K i\,p_i=K/H_K<(K+1)/2$, and the real-valued sample counts needed to hit the top $m$ with probability at least $1-\delta$ satisfy $n_{\mathrm{elite}}=\log\delta/\log(1-p_{\mathrm{elite}})<\log\delta/\log(1-p_{\mathrm{uniform}})=n_{\mathrm{uniform}}$.

\medskip\noindent\textit{Proof.} The top-$m$ probability under Eq.~(\ref{eq:elite}) is $\sum_{i=1}^m(1/i)/H_K=H_m/H_K$. Because $1/i$ is decreasing, the average of its first $m$ terms is larger than the average of its first $K$ terms when $m<K$, so $p_{\mathrm{elite}}>p_{\mathrm{uniform}}$. The expected rank is $\sum_{i=1}^K i(1/i)/H_K=K/H_K$, and the initial-segment inequality gives the stated comparison with uniform sampling. The probability of missing the top $m$ in $n$ independent samples is $(1-p_{\mathrm{elite}})^n$ under Eq.~(\ref{eq:elite}) and $(1-p_{\mathrm{uniform}})^n$ under uniform sampling; solving $1-(1-p)^n\ge1-\delta$ gives the displayed sample requirements, and $p_{\mathrm{elite}}>p_{\mathrm{uniform}}$ makes the elite-guided requirement smaller.
\hfill\rule{0.35em}{0.35em}

\medskip\noindent Within a fixed elite set, Eq.~(\ref{eq:elite}) therefore both concentrates reuse on highly ranked plans, code patterns, and execution feedback and reduces the number of independent samples needed to revisit a top-$m$ solution. 

\section{MLE-Bench Benchmark and Evaluation Metrics}
\label{appen:benchmark}

\subsection{MLE-Bench}
We evaluate \sname on MLE-Bench~\citep{mle-bench}, a benchmark introduced by OpenAI for assessing autonomous machine learning engineering. MLE-Bench comprises 75 carefully curated Kaggle competitions spanning natural language processing, computer vision, signal processing, and tabular data analysis. These competitions are selected from 586 candidates through manual screening by ML engineers, ensuring each task represents authentic and challenging ML engineering work. The dataset includes competitions of varying complexity: 22 low-complexity tasks (solvable by experienced engineers in under 2 hours), 38 medium-complexity tasks (2--10 hours), and 15 high-complexity tasks (over 10 hours), covering 15 distinct problem categories.

Each competition includes the original problem description, datasets with reconstructed train-test splits, local grading code, and human baseline performance from Kaggle leaderboards. This setup enables direct comparison between AI agents and human competitors while maintaining evaluation integrity. The benchmark employs medal achievement rates as the primary metric, where agents must reach bronze, silver, or gold medal thresholds based on their performance relative to human participants. Agents must work autonomously within time constraints (24-hour time limit) to produce valid submission files.

\subsection{Evaluation Metrics}
\label{appen:eval}
We use the following metrics to evaluate performance on MLE-Bench. All thresholds and percentile data are officially provided by Kaggle and MLE-Bench.

\begin{itemize}[leftmargin=*]
    \item \textbf{Medal Rate (All, in \%)}: the percentage of tasks on which the submission earns a medal (gold, silver, or bronze). We additionally report the medal rate stratified by task complexity (Low / Medium / High).
    \item \textbf{Gold Medal Rate (Gold, in \%)}: the percentage of tasks on which the submission earns a gold medal.
    \item \textbf{Valid Submission Rate (Valid, in \%)}: the percentage of tasks that produce a valid submission passing format and correctness checks.
    \item \textbf{Above Median Rate (Med+, in \%)}: the percentage of tasks on which the submission beats half of the human competitors.
    \item \textbf{Beat Ratio (in \%)}: the average percentage of human competitors whose performance is surpassed by the agent's submission.
\end{itemize}

\section{Hyperparameters}
\label{appen:hyper}

Table~\ref{tab:hyper} lists the key hyperparameters used in all \sname experiments. Values are kept fixed across all 75 MLE-Bench tasks unless otherwise stated.

\begin{table}[h]
\centering
\caption{Default hyperparameter configuration of \sname.}
\label{tab:hyper}

\setlength{\tabcolsep}{4pt}
\renewcommand{\arraystretch}{1.12}
\footnotesize

\begin{tabularx}{0.95\linewidth}{@{}lXr@{}}
\toprule
\textbf{Hyperparameter} & \textbf{Description} & \textbf{Default} \\
\midrule

\multicolumn{3}{@{}l}{\textit{General Search}}\\[-2pt]
\arrayrulecolor{black!30}\cmidrule(lr){1-2}\arrayrulecolor{black}
steps                      & Max search steps                                & 500 \\
time\_limit                & Total time limit per task                        & 12\,h \\
parallel\_search\_num      & Parallel branches                                & 3 \\
initial\_drafts            & Initial drafts                                   & 3 \\
max\_drafts                & Max branches from primary expansion               & 5 \\
max\_fusion\_drafts        & Max additional branches from aggregation           & 2 \\
temperature                & LLM decoding temperature                         & 1.0 \\

\midrule
\multicolumn{3}{@{}l}{\textit{Progressive MCGS}}\\[-2pt]
\arrayrulecolor{black!30}\cmidrule(lr){1-2}\arrayrulecolor{black}
exploration\_constant      & UCT exploration constant $c_0$                  & $\sqrt{2}$ \\
lower\_bound               & UCT lower bound $c_{\min}$                      & 0.5 \\
phase\_ratios              & UCT decay phase ratios                           & $(0.3, 0.7)$ \\
explore\_switch\_start     & Soft-switch start time ratio                     & 0.5 \\
explore\_switch\_end       & Soft-switch end time ratio                       & 0.7 \\
min\_exploration\_weight   & Minimum exploration weight $w_{\min}$           & 0.2 \\
elite\_topk                & Top-$K$ candidates in elite-guided exploitation  & 3 \\

\midrule
\multicolumn{3}{@{}l}{\textit{Stagnation Detection}}\\[-2pt]
\arrayrulecolor{black!30}\cmidrule(lr){1-2}\arrayrulecolor{black}
branch\_stagnation\_threshold  & Branch-level stagnation $\tau_{\text{branch}}$ & 3 \\
topk\_stagnation\_threshold    & Global-level stagnation $\tau_{\text{global}}$ & 6 \\

\midrule
\multicolumn{3}{@{}l}{\textit{Retrospective Memory}}\\[-2pt]
\arrayrulecolor{black!30}\cmidrule(lr){1-2}\arrayrulecolor{black}
memory\_similarity\_threshold  & Retrieval similarity threshold           & 0.7 \\
memory\_embedding\_model       & Sentence embedding model                 & BGE-base-en-v1.5 \\

\bottomrule
\end{tabularx}
\end{table}

\section{Detailed Component Analysis}
\label{appen:detailed_ablation}

To complement the component-level ablation in \S\ref{sec:ablation}, we evaluate individual mechanisms within Progressive MCGS and Retrospective Memory on nine MLE-Bench tasks spanning three complexity levels. The subset comprises three low-complexity tasks (siim-\allowbreak{}isic-\allowbreak{}melanoma-\allowbreak{}classification, text-\allowbreak{}normalization-\allowbreak{}challenge-\allowbreak{}russian-\allowbreak{}language, and the-\allowbreak{}icml-\allowbreak{}2013-\allowbreak{}whale-\allowbreak{}challenge-\allowbreak{}right-\allowbreak{}whale-\allowbreak{}redux), four medium-complexity tasks (cdiscount-\allowbreak{}image-\allowbreak{}classification-\allowbreak{}challenge, cassava-\allowbreak{}leaf-\allowbreak{}disease-\allowbreak{}classification, hubmap-\allowbreak{}kidney-\allowbreak{}segmentation, and tweet-\allowbreak{}sentiment-\allowbreak{}extraction), and two high-complexity tasks (nfl-\allowbreak{}player-\allowbreak{}contact-\allowbreak{}detection and 3d-\allowbreak{}object-\allowbreak{}detection-\allowbreak{}for-\allowbreak{}autonomous-\allowbreak{}vehicles). Each configuration disables one mechanism while keeping all other settings unchanged.

\begin{table}[t]
\centering
\caption{Detailed component analysis on 9 representative tasks. Each row disables one mechanism within Progressive MCGS or Retrospective Memory while keeping all others unchanged.}
\vspace{-0.5em}
\label{tab:fine_grained_ablation}
\small
\setlength{\tabcolsep}{6pt}
\renewcommand{\arraystretch}{1.05}
\begin{tabular}{lcc}
\toprule
\textbf{Configuration} & \textbf{Medal (\%)} & \textbf{Beat Ratio (\%)} \\
\midrule
\rowcolor[RGB]{222,236,215}
\textbf{\sname} & \textbf{66.67} & \textbf{82.43} \\
\midrule
\multicolumn{3}{l}{\emph{Progressive MCGS}} \\
\quad w/o Evolution      & 33.33 & 74.95 \\
\quad w/o Cross-branch   & 55.56 & 75.93 \\
\quad w/o Elite-Guided   & 55.56 & 71.39 \\
\midrule
\multicolumn{3}{l}{\emph{Retrospective Memory}} \\
\quad w/o Knowledge Base & 44.44 & 76.07 \\
\quad w/o Global Memory  & 44.44 & 73.58 \\
\bottomrule
\end{tabular}
\end{table}

As shown in Table~\ref{tab:fine_grained_ablation}, removing intra-branch evolution causes the largest medal-rate drop among the internal mechanisms of Progressive MCGS, from 66.67\% to 33.33\%. This indicates that reusing recent branch history is critical for preventing the agent from repeatedly making similar mistakes across iterations. Removing cross-branch reference and Elite-Guided exploitation leads to milder medal-rate decreases, but they affect different aspects of performance. Cross-branch reference provides external guidance from high-performing solutions discovered in other branches, helping the agent escape stagnant local trajectories. Elite-Guided exploitation mainly improves leaderboard ranking by further refining already competitive solutions toward higher-performing ones, as reflected by the lowest beat ratio after its removal.
For the memory system, removing either the Knowledge Base or Global Memory reduces the medal rate to 44.44\%, showing that both sources of experience contribute to performance. However, removing Global Memory leads to a lower beat ratio than removing the Knowledge Base, suggesting that dynamically accumulated experience has a stronger impact on overall solution quality during long-horizon search. The Knowledge Base mainly provides task-relevant priors for cold-start initialization, while Global Memory continuously accumulates and reuses task-specific experience throughout the search process.

\section{Detailed Results with Different LLMs}
\label{appen:multi_model}

To provide a detailed view of per-task performance across different LLM backbones, we report the scores of Gemini-3.1-Pro-preview, GPT-5.5, DeepSeek-v4-Pro, and Kimi-K2.6 on 8 representative MLE-Bench tasks covering Image, NLP, and Audio domains. As shown in Table~\ref{tab:multi_model_score}, each model has its own strengths across different tasks and domains, with no single model dominating all tasks. All four backbones produce competitive results under the same \sname pipeline, confirming that the framework is not tied to a specific LLM.

\begin{table*}[t]
    \centering
    \caption{Score comparison of \sname across four LLM backbones on 8 representative MLE-Bench tasks. Best result for each task is highlighted in \textbf{bold}.}
    \label{tab:multi_model_score}
    \resizebox{1.0\linewidth}{!}{
    \setlength{\tabcolsep}{6pt}
    \begin{tabular}{llcccc}
      \toprule
      \textbf{Task} & \textbf{Metric} & \textbf{Gemini-3.1-Pro-preview} & \textbf{GPT-5.5} & \textbf{DeepSeek-v4-Pro} & \textbf{Kimi-K2.6} \\
      \cmidrule(lr){3-6}
      \multicolumn{6}{l}{\textit{Image Tasks}}\\
      \midrule
      cassava-leaf-disease-classification      & Accuracy $\uparrow$   & 0.8984 & 0.8999 & \textbf{0.9032} & 0.8905 \\
      ranzcr-clip-catheter-line-classification & AUC $\uparrow$        & \textbf{0.9638} & 0.9568 & 0.9614 & 0.8880 \\
      siim-isic-melanoma-classification        & AUC $\uparrow$        & 0.9252 & 0.9045 & \textbf{0.9294} & \textbf{0.9294} \\
      \midrule
      \multicolumn{6}{l}{\textit{NLP Tasks}}\\
      \midrule
      tweet-sentiment-extraction               & Jaccard $\uparrow$    & 0.7136 & \textbf{0.7216} & 0.7136 & 0.7195 \\
      spooky-author-identification             & Logloss $\downarrow$  & \textbf{0.2175} & 0.2324 & 0.2189 & 0.2305 \\
      random-acts-of-pizza                     & AUC $\uparrow$        & 0.6992 & \textbf{0.7888} & 0.7782 & 0.7649 \\
      \midrule
      \multicolumn{6}{l}{\textit{Audio Tasks}}\\
      \midrule
      the-icml-2013-whale-challenge-right-whale-redux & AUC $\uparrow$ & \textbf{0.9947} & \textbf{0.9947} & 0.9938 & 0.9934 \\
      freesound-audio-tagging-2019             & lwlrap $\uparrow$     & 0.7400 & 0.6766 & 0.6838 & \textbf{0.7586} \\
      \bottomrule
    \end{tabular}
    }
\end{table*}

\section{Case Study}
\label{appen:case}

We present representative cases illustrating the three graph-based expansion operators. Each case is drawn from an actual run on MLE-Bench.

\subsection{Intra-branch Evolution}
\label{appen:case_evo}

\begin{figure}[h]
\centering
\includegraphics[width=\linewidth]{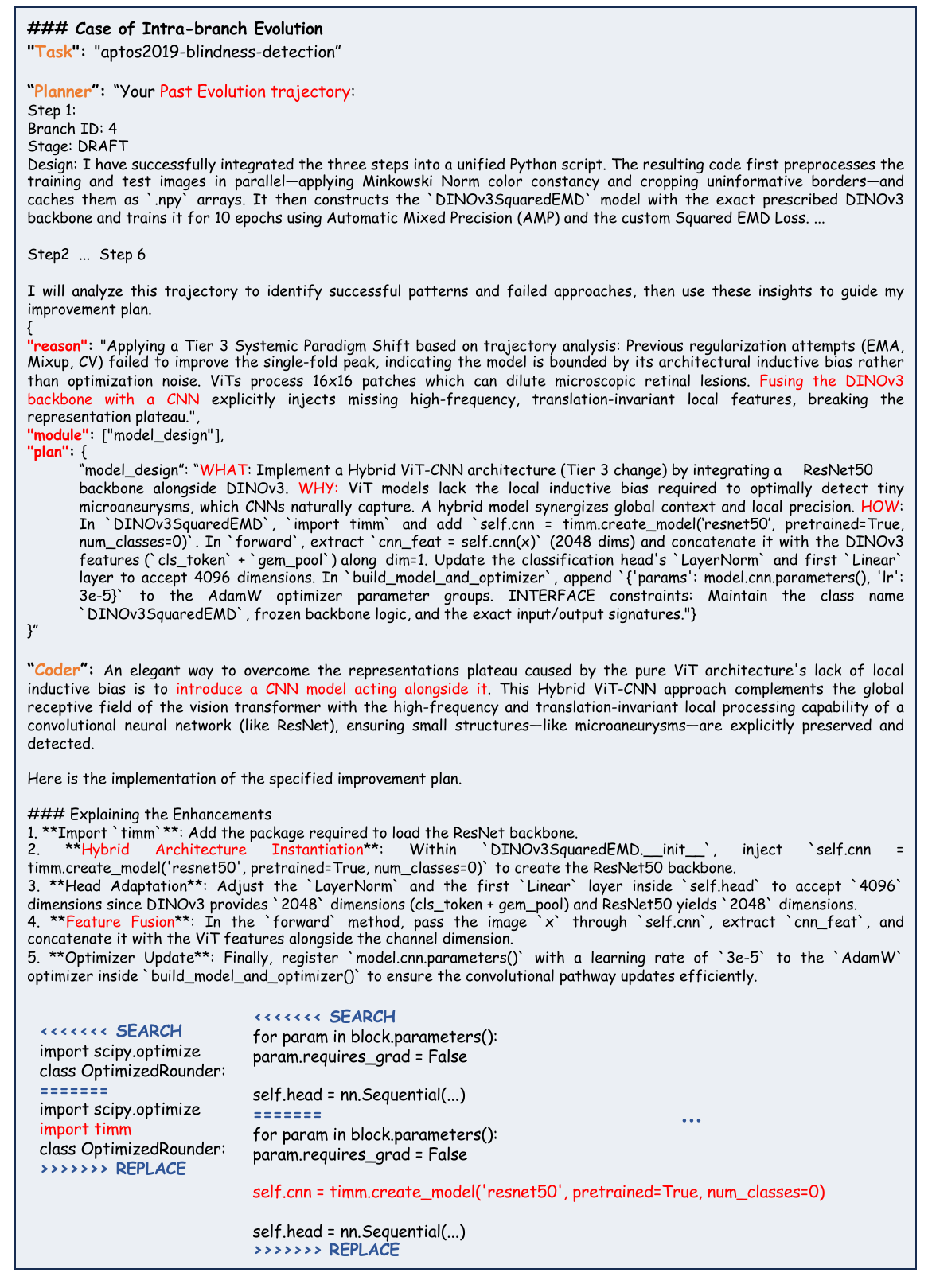}
\caption{An intra-branch evolution case on the \texttt{aptos2019-blindness-detection} task.}
\label{fig:case_evo}
\end{figure}

Figure~\ref{fig:case_evo} shows an intra-branch evolution case on the \texttt{aptos2019-blindness-detection} task. After six successive draft and improvement steps, the branch reaches a plateau where prior regularization attempts (EMA, Mixup, cross-validation) all failed to improve the score. The Evolution Agent reviews the local trajectory, identifies the bottleneck as architectural, and proposes fusing a DINOv3 backbone with a ResNet50. The coder implements the change via diff-mode edits.

\subsection{Cross-branch Reference}
\label{appen:case_ref}

\begin{figure}[h]
\centering
\includegraphics[width=\linewidth]{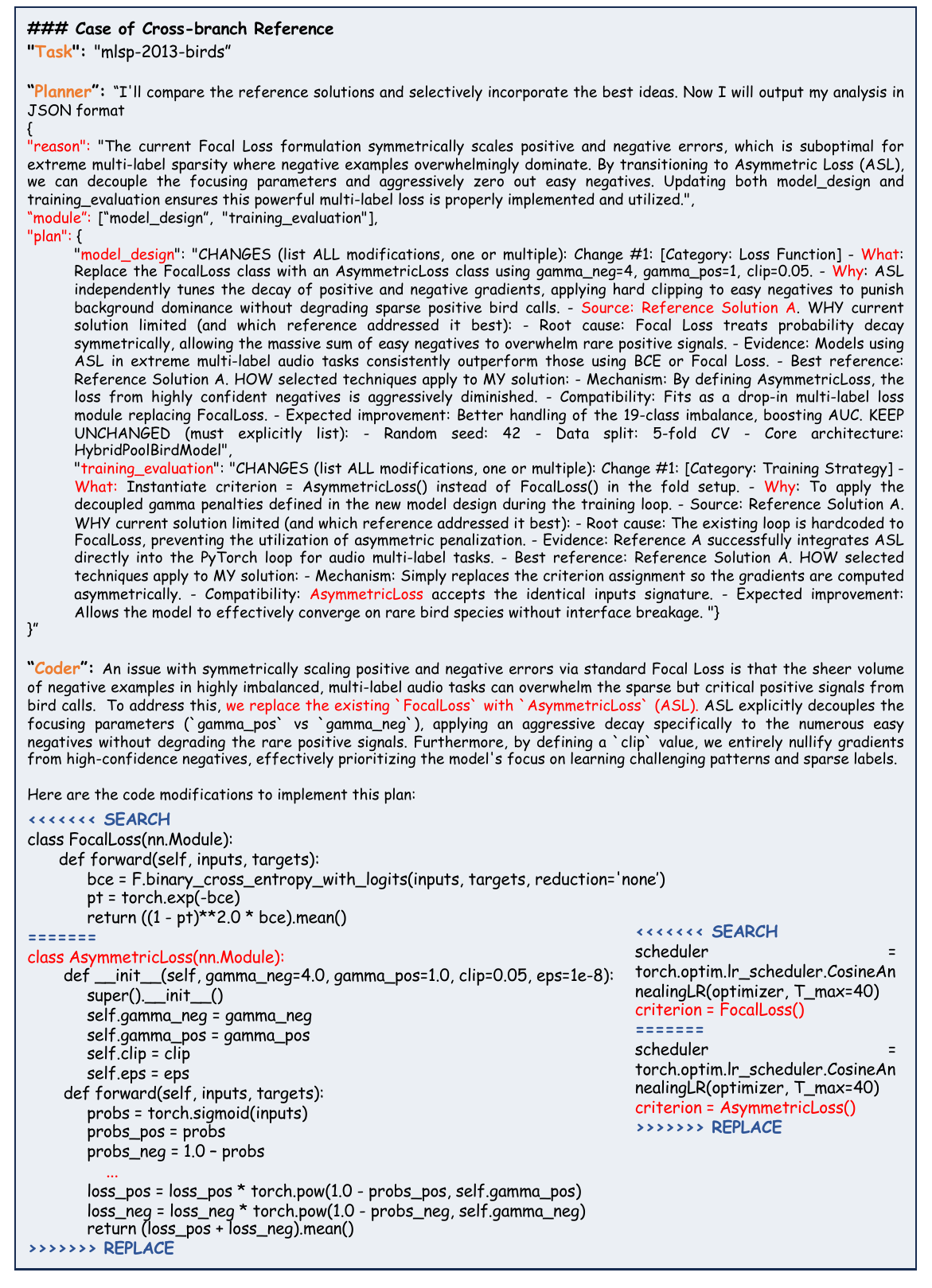}
\caption{A cross-branch reference case on the \texttt{mlsp-2013-birds} task.}
\label{fig:case_ref}
\end{figure}

Figure~\ref{fig:case_ref} shows a cross-branch reference case on the \texttt{mlsp-2013-birds} task. When the current branch stagnates with a symmetric Focal Loss, the Fusion Agent identifies an alternative loss design (Asymmetric Loss) from a strong solution in another branch. The coder replaces \texttt{FocalLoss} with \texttt{AsymmetricLoss} via diff-mode edits.

\subsection{Multi-branch Aggregation}
\label{appen:case_agg}

\begin{figure}[h]
\centering
\includegraphics[width=\linewidth]{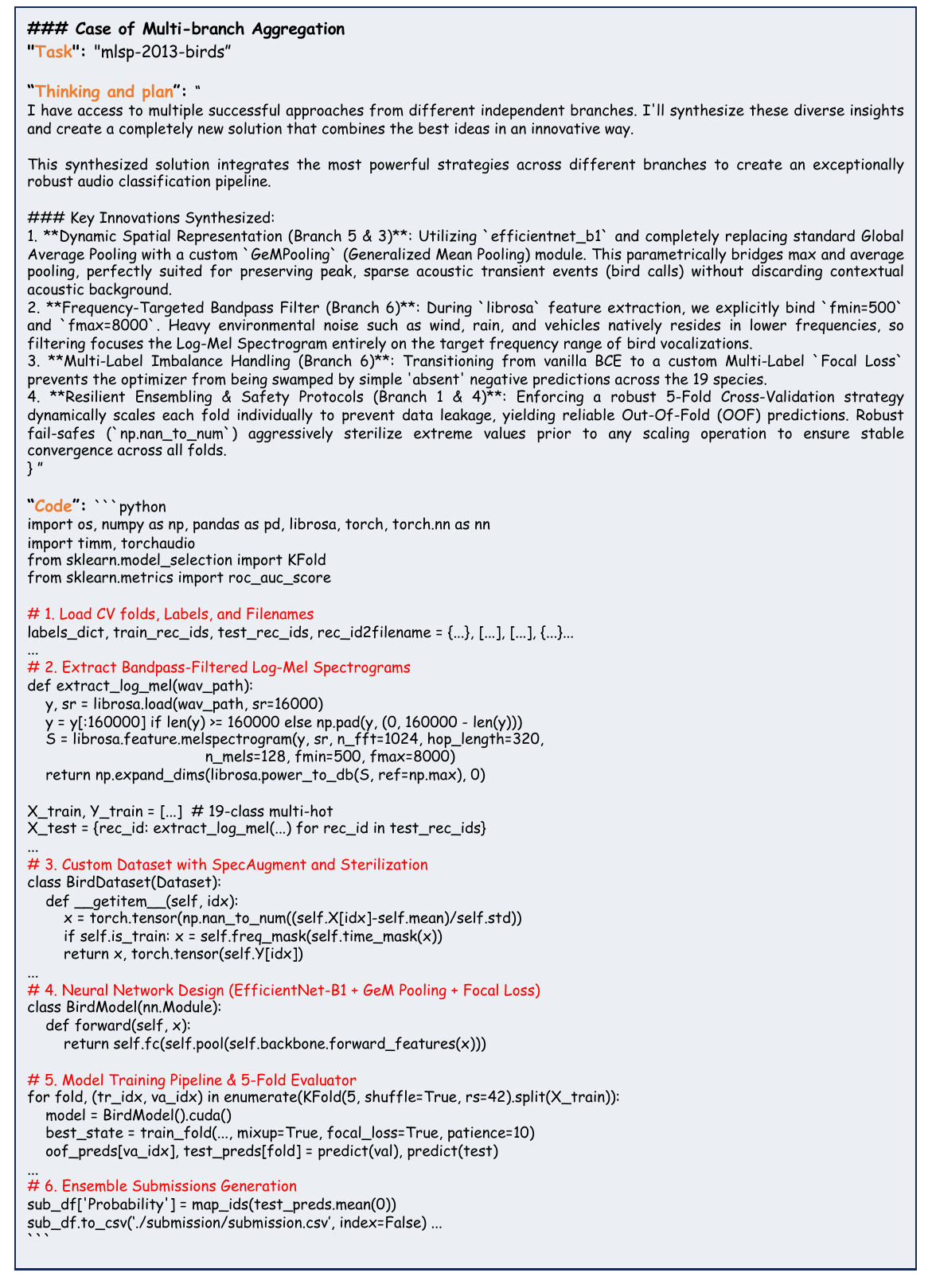}
\caption{A multi-branch aggregation case on the \texttt{mlsp-2013-birds} task.}
\label{fig:case_agg}
\end{figure}

Figure~\ref{fig:case_agg} shows a multi-branch aggregation case on the same \texttt{mlsp-2013-birds} task. After global stagnation is detected, the Aggregation Agent synthesizes successful components from multiple branches (EfficientNet-B1 with GeM pooling, bandpass filter, Multi-Label Focal Loss, 5-fold cross-validation) into a new branch starting point.


\end{document}